# Incorporating Domain Knowledge in Matching Problems via Harmonic Analysis

**Deepti Pachauri**[†], **Maxwell Collins**[†], **Risi Kondor**[§], **Vikas Singh**[‡†]

[†]Dept. of Computer Sciences, [‡]Dept. of Biostatistics & Med. Informatics, University of Wisconsin Madison.
[§]Dept. of Computer Science and Dept. of Statistics, University of Chicago.
{pachauri,mcollins}@cs.wisc.edu    risi@uchicago.edu    vsingh@biostat.wisc.edu

## Abstract

Matching one set of objects to another is a ubiquitous task in machine learning and computer vision that often reduces to some form of the quadratic assignment problem (QAP). The QAP is known to be notoriously hard, both in theory and in practice. Here, we investigate if this difficulty can be mitigated when some additional piece of information is available: (a) that all QAP instances of interest come from the same application, and (b) the correct solution for a set of such QAP instances is given. We propose a new approach to accelerate the solution of QAPs based on *learning* parameters for a modified objective function from prior QAP instances. A key feature of our approach is that it takes advantage of the algebraic structure of permutations, in conjunction with special methods for optimizing functions over the symmetric group $\mathbb{S}_n$ in Fourier space. Experiments show that in practical domains the new method can outperform existing approaches.

## 1. Introduction

Matching one set of objects to another is a fundamental problem in computer science. In computer vision it arises in the context of finding the correspondence between multiple images of the same scene taken from different viewpoints (Fig. 1). In bioinformatics one must align sequences of genes and amino acids. In machine learning one often needs to align examples before a meaningful similarity measure can be computed between them (Pachauri et al., 2011). Some of these problems reduce to a linear assignment problem

$$\arg\max_{\sigma \in \mathbb{S}_n} \sum_{i=1}^{n} Q_{i,\sigma(i)}, \qquad (1)$$

where $\mathbb{S}_n$ is the set of all permutations of $\{1, 2, \ldots, n\}$, and $Q_{ij}$ captures the negative cost of matching object $x_i$ to object $y_j$. This problem can be solved in $O(n^3)$ time using the well-known Kuhn–Munkries (or Hungarian) algorithm.

The limitation of (1) is that it does not take into account the relationships of the $x_i$'s (and the $y_i$'s) to each other. For example, in matching feature points (or landmarks) between images, not only do we want landmarks in one image to be matched to similar landmarks in the other image, but we also want the distances between landmarks in the first image to be similar to the distances between the corresponding landmarks in the second one. This leads to a more general optimization problem

$$\arg\max_{\sigma \in \mathbb{S}_n} \sum_{i,j=1}^{n} A_{\sigma(i),\,\sigma(j)}\, A'_{i,j}, \qquad (2)$$

known as the **quadratic assignment problem** (QAP). One way to think of (2) is to view it as the problem of matching two weighted graphs with adjacency matrices $A$ and $A'$ so as to maximize the overlap between them.

Unfortunately, the QAP is NP–hard, and it is also notoriously hard to approximate or solve heuristically. Combinatorial search methods such as branch and bound almost never manage to solve real world QAP instances in polynomial time, while convex optimization methods are thwarted by the fact that the feasible region (called the second order permutation polytope) has an exponential number of faces. From a purely empirical point of view the most successful approaches are ant-methods and the like, which are very heuristic algorithms with no optimality guarantees whatsoever.

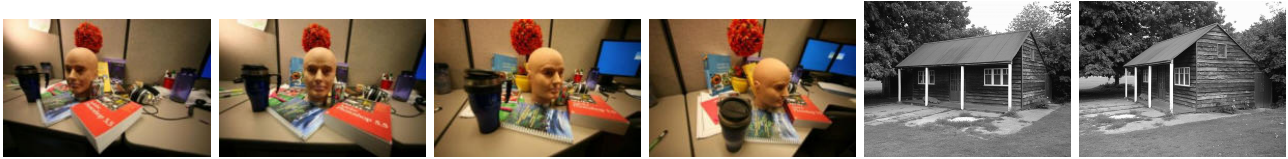

*Figure 1.* A set of images of the same object. Assuming that the images are not related by a simple transformation such as homography (Hartley & Zisserman, 2000), the matching seeks to assign feature points (nodes) in one image to its most similar feature point in the other image. Typically, there is also a cost associated with alignment of corresponding edges.

In a fraction of the cases they find a near-optimal solution very fast, while in many other cases they fail, and there is no way of knowing whether we are in the former domain or the latter.

It is natural to ask whether one can use side information to make QAP easier. Here, we propose a new approach for doing this by *learning* a modified objective function $f^\omega$ from a set of prior "training" QAP instances. The two criteria that $f^\omega$ must satisfy are

1. $\arg\max_{\sigma \in \mathbb{S}_n} f^\omega(\sigma)$ should be close to the maximum of the original objective function $f$.
2. $f^\omega$ should be much easier to optimize than $f$.

The vector $\boldsymbol{\omega}$ that parametrizes $f^\omega$ is determined by methods similar to Structural Risk Minimization. This form of risk minimization strategy (Finley & Joachims, 2008), as in Structural SVMs, when applied to the training set, allows our parameter $\boldsymbol{\omega}$ to generalize to unseen examples well. Note that Structural SVMs are very well understood if the original inference problem ($\arg\max f^\omega$) is poly-time solvable (see (Tsochantaridis et al., 2006; Taskar et al., 2003)). Unfortunately, the original graph matching problem requires finding a $\boldsymbol{\omega}$ such that the minimizer of $f^\omega(\cdot)$ matches $\sigma^*$ — this is itself a QAP. Existing theoretical guarantees for Structural SVM are known to be far less satisfactory for such intractable problems (Tsochantaridis et al., 2006; Joachims et al., 2009).

Fortunately, we may observe that the $\sigma$'s of our interest are not arbitrary objects, rather constitute the symmetric group, and this opens the door to look at the properties of specific sub-classes of $f^\omega$ and/or $\sigma$ that may provide useful insights into efficient solution strategies. Specifically, this view allows us to leverage an entire spectrum of tools from abstract algebra including non-commutative harmonic analysis and fast Fourier transforms on groups to develop efficient algorithms (Rudin, 1962; Diaconis, 1988). The real promise of this new approach lies in its generality: by setting matching problems in a broader algebraic framework it has the potential to serve as a basis for developing new matching algorithms that exploit the representation theory of groups and are better informed by the the characteristics of the underlying vision or learning task.

While the model developed here is applicable to the learning version of most of the above scenarios, to make our exposition concrete, we will restrict our attention to the problem of learning parameters for graph matching. The Learning Graph Matching problem (Caetano et al., 2009) seeks to solve for parameters of compatibility functions so that the solution from an approximate solution matches the permutation matrix provided by the user as best as possible. Since the main objects of interest are permutation matrices, it makes this an ideal sandbox to develop and present our main ideas.

**Graph Matching and other Related Work.**
Graph Matching is the problem of finding correspondences between the nodes of two graphs to maximize alignment. One way to model it is to write it as a Quadratic Assignment Problem (QAP), which is among the most well-studied combinatorial optimization problems in the literature (Pardalos et al., 1994; Cela, 1998; Umeyama, 1988; Vishwanathan et al., 2007). Many alternative approaches for graph matching are also known (Leordeanu & Hebert, 2005; Hancock & Wilson, 2009; Caelli & Caetano, 2005). In the context of *learning* graph matching (Caetano et al., 2009; Leordeanu & Hebert, 2009), one is interested in the following question: if the optimal correspondence between the nodes of a pair of graphs is known (and many such pairs are available), how should one use this knowledge to learn correspondences between another pair of graphs which were extracted under similar *conditions*. The notion of *conditions* reflects properties of the application under study – for instance, (Caetano et al., 2009) uses the example of image pairs acquired under similar illumination from an airport surveillance camera, where the matching task refers to aligning the "feature points" from such images. The to be determined parameter $\boldsymbol{\omega}$ then corresponds to weights which appropriately adjust the joint feature map of node and edge compatibilities, so that the match found by the solver agrees with the user provided solution. Learning graph matching serves another very important need – by restricting our focus to a *specific* domain and tuning weights that best reflect practical considerations in that application, a less sophisticated approach may still be able to obtain good quality solutions in a fixed amount of time or memory (Xu et al., 2007). This has implications in most situations where training data is available. The algorithm in (Caetano et al., 2009) uses a nice structure learning formalization for this problem. But finding the most violated constraint precisely in the construction (Caetano et al., 2009) itself involves solving a QAP, and can only be approximately estimated (e.g., via its linear assignment relaxation).

**This Paper.** We recast the learning problem above using ideas from the theory of non-commutative Fourier analysis on the symmetric group. Our core algorithm is defined on a certain tree which has the permutations ($\sigma \in \mathbb{S}_n$) at the leaves and each branch corresponds to a coset of $\mathbb{S}_n$; the modulation of the edge and node compatibility functions then consists of solving a set of simple convex optimization models defined in Fourier space. Broadly, the approach performs *stochastic descent*-like weight updates to induce a margin between the path that leads to the correct user-provided $\sigma^*$, relative to all other $\sigma$'s, for each training example. The **contribution** of this paper is the parameter learning framework for a class of combinatorial problems where the solution is a candidate in the symmetric group $\mathbb{S}_n$. We show how the representation theory of $\mathbb{S}_n$ makes the procedure computationally tractable, and how Branch and Bound schemes can be modified to learn information relevant for problem instances coming from an application of interest.

## 2. The form of QAP in Fourier space

The key aspect of the QAP which our approach exploits is that $\mathbb{S}_n$, the set of permutations that we need to optimize over, is a *group*, called the symmetric group over $n$ letters. Recall that this means, with respect to the natural notion of multiplication, which is $(\sigma_2 \sigma_1)(i) = \sigma_2(\sigma_1(i))$, permutations in $\mathbb{S}_n$ satisfy the following axioms:

1. $\sigma_2 \sigma_1 \in \mathbb{S}_n$ (closure);
2. $\sigma_3(\sigma_2 \sigma_1) = (\sigma_3 \sigma_2)\sigma_1$ (associativity);
3. there is an identity $e \in \mathbb{S}_n$ such that $\sigma_1 e = \sigma_1$;
4. every $\sigma$ has an inverse $\sigma^{-1} \in \mathbb{S}_n$ such that $\sigma^{-1}\sigma = \sigma\sigma^{-1} = e$

Remarkably, these axioms are sufficient to define a meaningful notion of harmonic analysis on $\mathbb{S}_n$. However, because $\mathbb{S}_n$ is non-commutative ($\sigma_2 \sigma_1 \neq \sigma_1 \sigma_2$), the Fourier components will be matrices. In particular, the Fourier transform of a general function $f: \mathbb{S}_n \to \mathbb{R}$ consists of the matrices

$$\widehat{f}(\lambda) = \sum_{\sigma \in \mathbb{S}_n} f(\sigma)\,\rho_\lambda(\sigma), \qquad (3)$$

where $\lambda = (\lambda_1, \lambda_2, \ldots, \lambda_k)$ is the so-called integer partition of $n$ and plays the role of "frequency", while $\rho_\lambda: \mathbb{S}_n \to \mathbb{C}^{d_\lambda \times d_\lambda}$ is the corresponding irreducible representation (irrep) of $\mathbb{S}_n$. For more details on these algebraic concepts, see this paper's extended version.

Kondor (2010) recently showed that the machinery of non-commutative Fourier transforms can be used to advantage in solving the QAP. One key observation is that if one defines so-called graph functions $f_A: \mathbb{S}_n \to \mathbb{R}$,

$$f_A(\sigma) = A_{\sigma(n), \sigma(n-1)},$$

then the objective function of the QAP,

$$f(\sigma) = \sum_{i,j=1}^{n} A_{\sigma(i),\,\sigma(j)}\, A'_{i,j} \qquad (4)$$

can easily be expressed in Fourier space as

$$\widehat{f}(\lambda) = \frac{1}{(n-2)!}\widehat{f}_A(\lambda)\widehat{f}_{A'}(\lambda)^\dagger,$$

where $^\dagger$ stands for the conjugate transpose. Given that functions of the form (4) are band-limited to $\{(n), (n-1,1), (n-2,2), (n-2,1,1)\}$ (i.e., all components of $\widehat{f}_A$ other than those indexed by these integer partitions are identically zero), the Fourier transform of the objective function will also be restricted to $\widehat{f}((n)), \widehat{f}((n-1,1)), \widehat{f}((n-2,2))$ and $\widehat{f}((n-2,1,1))$. In fact, this observation has already been made in (Rockmore et al., 2002a). We will use this idea for computing QAP bounds to be discussed shortly.

Another key ingredient in our proposed approach is the technology of Fast Fourier Transforms (FFTs) on $\mathbb{S}_n$, going back to the work of Clausen (1989). Recall that given a subgroup $H$ of $\mathbb{S}_n$ and a permutation $\sigma \in \mathbb{S}_n$, the set $\sigma H = \{\,\sigma\tau \mid \tau \in H\,\}$ is called a left $H$–coset. Similarly, one can talk about right $H$–cosets such as $H\sigma = \{\,\tau\sigma \mid \tau \in H\,\}$, and two-sided cosets such as $\sigma_1 H \sigma_2 = \{\,\sigma_1 \tau \sigma_2 \mid \tau \in H\,\}$. FFTs generally work by first computing Fourier transforms of $f$ restricted to small cosets, and then recursively assembling such small transforms into ever large ones until we reach the Fourier transform in (3) on the entire group. A QAP solver can use this structure to search $\mathbb{S}_n$ by employing the Inverse Fast Fourier Transform (iFFT) to restrict $f$ to various cosets and bound it. We illustrate a Coset tree in Figure 2 for $n=3$.

As briefly discussed above, the final component of Fourier space QAP solver are the bounds for $f$ (re-

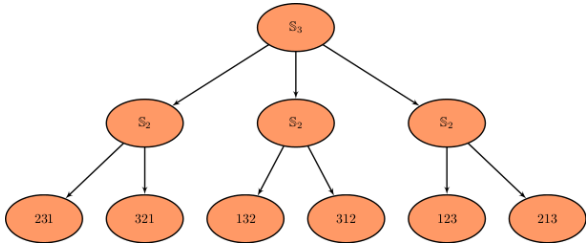

Figure 2. Left coset tree for $\mathbb{S}_3$ showing all members of $\mathbb{S}_3$ as leaves. Second level of the tree represents $\mathbb{S}_2$–coset of $\mathbb{S}_3$ which corresponds to candidate permutations $[\![1,3]\!], [\![2,3]\!], [\![3,3]\!]$.

stricted to various cosets) based on its Fourier components. The simplest such bound is

$$f(\sigma) \leq \frac{1}{n!} \sum_{\lambda} d_\lambda \left\| \widehat{f}(\lambda) \right\|_* \quad (5)$$

where $\|\ \|_*$ denotes the nuclear norm. Combining these components leads to a Branch and Bound type optimization algorithm that runs in $O(n^3)$ time per branch visited and is competitive with more conventional exact QAP solvers (Kondor, 2010). Unfortunately, in many practical problems the resulting algorithm still takes an exponential amount of time to run, simply because it needs to visit so many branches.

## 3. The Objective function for Learning

Observe that QAPs are hard because the objective function is relatively flat, and according to most reasonable metrics, the diameter of $\mathbb{S}_n$ is small compared to its size. In practical problems, however, one can sometimes overcome such seemingly insurmountable barriers by making inventive use of side information. In particular, we take the approach of using similar QAP instances (derived from the application of interest) to learn a modified objective function that will more effectively drive our algorithm to the correct solution. Since solving a QAP from scratch for each modification of parameters is clearly an intractable option, our main goal will be to adapt tools from non-commutative harmonic analysis to recast the problem in a way that sidesteps this burden. The framework described next formalizes this idea.

We first write the QAP objective function for Graph matching. The function is defined on the adjacency matrices of a graph pair, and has a band-limited structure observed in (Rockmore et al., 2002b). Next, we incorporate (i.e., parameterize) domain information within the objective. This will set us up to present the framework for learning these parameters given multiple instances of *related* graphs.

Given two graphs $G, G'$ of $n$ vertices with adjacency matrices $A, A'$, the standard *quadratic assignment problem* finds the permutation which best aligns the graph. We choose $\sigma$ which optimizes

$$\max_{\sigma \in \mathbb{S}_n} f(\sigma) = \sum_{i,j=1}^n A_{i,j} A'_{\sigma(i),\sigma(j)} \quad (6)$$

In the unweighted case, $f(\sigma)$ simply counts the number of edges which appear in both $G$ and the permuted graph $\sigma(G')$. The objective function of standard QAP is expressed as the **graph correlation** (Kondor, 2010) between the two graphs $G$ and $G'$

$$f(\sigma) = \frac{1}{(n-2)!} \sum_{\pi \in \mathbb{S}_n} f_A(\sigma\pi) f_{A'}(\pi) \quad (7)$$

where $f_A$ and $f_{A'}$ are defined in section 2. One may then systematically search the coset tree (2) via the standard Branch and Bound to maximize (7).

### 3.1. Parameterizing the QAP objective

QAP instances derived from Vision problems typically rely on some analytic form approximating the perceptual similarity between two feature points. For instance, how similar are a pair of shape features (Belongie et al., 2002), $u$ and $v$, extracted from different images? If a node to node match which is perceptually correct turns out to be only marginally better than many other incorrect matches, this necessarily suggests that the features are not very discriminative. One practical consequence is that a Branch and Bound type search will need to work much harder (i.e., explore many sub-trees) to find the global solution.

The core strategy is to incorporate domain information in the QAP objective. To do this, we use the simple idea of composing various QAPs to write a base QAP objective which has certain desirable properties. Since features are available in most applications, it seems logical to use this additional information in the design of a function that can be used to inform the base QAP objective. While this function might not be perfect due to the noisy features, standard learning algorithms can be used to learn shared structure by observing multiple instances drawn from the specific domain.

Using training data (examples from the application of interest) we will be able to *learn* the parameters such that learnt parameters will induce domain friendly QAP instances – biasing the search towards the more interesting matches first, while simultaneously suppressing the influence of misleading features.

## 3.2. Learning Graph Matching

There are various ways of extracting features in Vision. Each of them model the neighborhood relationship between the vertices of $G$ based on *similarity of features*. The list may include *edge features such as a Delaunay triangulation*, Euclidean distance between interest points, Shape context features etc. Consider $D$ such representations are at our disposal, where each encodes the corresponding graph neighborhood.

We use each of these representations to generate $D$ adjacency matrices for each graph. Each entry $A_{i,j}^d$ is a squared distance between the feature values of vertex $i$ and $j$ of $G$ according to representation $d \in D$. Similarly, we define $A'^d$ for $G'$. We can write QAP objective (7) as before using $(A^d, A'^d)$

$$f^d(\sigma) = \frac{1}{(n-2)!} \sum_{\pi \in \mathbb{S}_n} f_{A^d}(\sigma\pi) f_{A'^d}(\pi)$$

We want to find a match such that edge $(i,j)$ in $G$ should be assigned to an edge $(i',j')$ in $G'$ that is of a similar length (or weight) simultaneously in all encoded adjacency matrices. However, features are noisy and not very discriminative, and can lead to wrong assignments. The strategy is to parameterize each $f^d(\sigma)$ and write a parameterized QAP objective for learning

$$f^\omega(\sigma) = \sum_{d=1}^{D} f_{\omega_d}^d(\sigma) \quad (8)$$

where the subscript $\omega_d$ represents parameterization, that is, we use the parameter vector $\omega \in \mathbb{R}^D$ to modulate the QAP function $f^\omega(\sigma)$.

The learning algorithm then essentially amounts to adjusting the $\omega$ appearing in the *parameterized* QAP objective using the *true* assignments $\sigma^*$ given for each training pair $(G_m, G'_m)$. Our general goal is to find

$$\omega^* = \arg\min_\omega \sum_{m=1}^{M} L(\hat{\sigma}_m(\omega), \sigma_m^*) + \Omega(\omega) \quad (9)$$

for some loss function $L(\cdot, \cdot)$ and regularizer $\Omega$, where $\hat{\sigma}_m(\omega)$ is the optimal permutation for example $m$. Note that $\hat{\sigma}_m(\omega)$ itself corresponds to solving a QAP objective given a QAP objective modulated by parameter $\omega$. The goal is to "learn" $\omega$ by performing *stochastic descent* at *each level of a tree of cosets*. In our experiments, we implemented a 0/1 loss. We observe that there may be many $\omega$'s that will yield the same 0/1 loss; the regularization $\Omega$ used here seeks the minimal difference from the $\omega$ in the previous step, so that the path/node leading to $\sigma_m^*$ is preferable to any other node at this level, by a small margin. The exact algorithm is discussed next.

## 4. Algorithm: Learning in Fourier space

In the following, we explain our Fourier space bounds for learning. The Fourier space machinery discussed in §2 bounds the objective function for learning (introduced in §3). More details on FFT and irreducible representation of $\mathbb{S}_n$ are in this paper's longer version.

### 4.1. Fourier Space Upper Bounds for Learning

When working in a so-called adapted system of representations, the Fourier matrices at level $n$ of a general function $f$ can be expressed in terms of the Fourier matrices at level $n-1$ as

$$\widehat{f}(\lambda) = \sum_{i=1}^{n} \rho_\lambda(\llbracket i, n \rrbracket) \bigoplus_{\mu \in \lambda\downarrow_{n-1}} \widehat{f_i}(\mu). \quad (10)$$

where $\bigoplus$ is the direct sum of Fourier matrices, and $\lambda\downarrow_{n-1}$ is used to denote the integer partitions of the "ancestor partitions".

The Fourier space QAP solver proceeds by searching the tree of cosets, which corresponds to assigning vertex $n, n-1, \ldots$ of $G$ to some sequence of vertices of $G'$. At level $n-k$ in the coset tree, it decides which vertex of $G'$ it should assign the vertex $n-k$ to by comparing the Fourier space bounds. Kondor (2010) used the inverse map of (10) to define the Fourier space bounds for standard QAP, where the inverse map at level $n-1$ is given by

$$\hat{f}_i(\mu) = \sum_{\lambda \in \mu\uparrow^n} \frac{d_\lambda}{nd_\mu} [\rho_\lambda(\llbracket i, n \rrbracket)^\top \hat{f}(\lambda)]_\mu, \quad (11)$$

Fourier space bounds are defined as

$$B_{n \to i} = \sum_{\mu \vdash n-1} \| \sum_{\lambda \in \mu\uparrow^{n-1}} \frac{d_\lambda}{nd_\mu} [\rho_\lambda(\llbracket i, n \rrbracket)^\dagger \hat{f}(\lambda)]_\mu \|_*, \quad (12)$$

$\hat{f}$ is replaced by $\hat{f}^\omega$ in (12) for learning correct parameters. It turns out that for a fixed $\omega$, $f^\omega(\sigma)$ is easily computable from each $f_{\omega_d}^d$ *entirely in Fourier space*, without having to perform a very costly full Fourier transform. The exact form in which $\omega$ interacts with $f_{\omega_d}^d$ depends on the problem formulation. To keep the presentation simple, we formulate our QAP objective as follows

$$f^\omega(\sigma) = \sum_{d=1}^{D} \omega_d f^d(\sigma) \quad (13)$$

The Fourier transform of $f^\omega$ can be expressed as

$$\widehat{f^\omega}(\lambda) = \sum_{d=1}^{D} \omega_d \widehat{f^d}(\lambda), \quad \lambda \vdash n \quad (14)$$

The fact that makes $f^\omega$ *learnable* is that the inverse map $\hat{f}_i^\omega(\mu)$ can be expressed as:

$\widehat{f}_i^\omega(\mu) = \sum_{d=1}^D \omega_d \widehat{f}_i^d(\mu)$, where $\widehat{f}_i^d(\mu)$ is the inverse map of $\widehat{f}^d(\lambda)$ according to (11). The identity follows directly from linearity of Fourier transform.

Ultimately, we may use the convex nature of the nuclear norm that makes *Jensen's inequality* handy to derive an easy-to-optimize set of bounds,

$$\|\widehat{f}_i^\omega(\mu)\|_* = \|\sum_{d=1}^D \omega_d \widehat{f}_i^d(\mu)\|_* \le \sum_{d=1}^D \omega_d \|\widehat{f}_i^d(\mu)\|_*. \quad (15)$$

With these concepts in hand, the only missing ingredient is the actual procedure to calculate the parameters $\omega$ that is presented next.

### 4.2. Stochastic gradient descent solver

In general, our loss function $L$ as in (9) is a hinge loss on the relative bounds between the correct nodes and their incorrect siblings. This takes the form

$$\sum_{k=1}^n \sum_{i \in \text{children}((n-k+1)^*)} \left[\widehat{f}_i^\omega(\mu) - \widehat{f}_{i_{n-k}^*}^\omega(\mu) + 1\right]^+ \quad (16)$$

which is summed over all examples, and along with the regularization represents the function we seek to minimize. And $i_{n-k}^*$ denotes the correct node at level $n-k$ for some $\sigma^*$. Note that bounding the siblings of the correct nodes will bound $f$ for *all* incorrect permutations.

We employ a stochastic gradient descent approach similar to (Shalev-Shwartz et al., 2007) on upper bounds defined in (12). During the training phase, we know which node to node assignment we should make, that is, we know which of these bounds, say the one corresponding to vertex $i_{n-k}^*$, we want to be the largest. A random training example and node $i$ is selected and compared to its correct sibling $i_{n-k}^*$. We then reduce the objective for one term by taking a gradient step on $\omega$. When $\Omega(w) = \frac{\nu}{2}\|\omega\|_2^2$ and $\widehat{f}_i^\omega$ is replaced with the bounds in (15), each update takes the form

$$\omega_d \leftarrow \omega_d - \eta \begin{cases} \|\widehat{f}_i^d(\mu)\|_* - \|\widehat{f}_{i_{n-k}^*}^d(\mu)\|_* + \frac{\nu}{M\mathcal{O}(n^2)}\omega_d \\ \frac{\nu}{M\mathcal{O}(n^2)}\omega_d \end{cases} \quad (17)$$

$\eta$ is an exponentially decaying step length parameter, $M$ is the number of training examples, and the $\frac{\nu}{M\mathcal{O}(n^2)}$ term arises from splitting the regularizer over all nodes considered by the optimization.

This is like structural SVM, where we try to find parameters so that the model predicts $i_{n-k}^*$ instead of $i$, with the bound for each correct node in the coset trees of the training set greater than its incorrect siblings by some margin. As per (15), learnt parameters are goodness measures of individual graph correlation function $f^d$ contributing in (13).

## 5. Experiments

To demonstrate proof of principle results of the proposed algorithm, we learnt parameters for compatibility functions in the context of graph matching. This has numerous applications, but the setting below considers the task of aligning two 2D images using local features extracted from the image data including interest points, and shape context features.

**Edge–features:** We performed Delaunay triangulation on interest points to generate unweighted edges. This provides unweighted adjacency matrix.
**Distance–features:** We used 2D coordinates of interest points to calculate Euclidean distance between points. We extracted weighted adjacency matrices at various scales using 2D distances.
**Shape Context–features:** As in (Caetano et al., 2009), we also included *Shape context* features (Belongie et al., 2002). Briefly, the feature vector is a descriptor in Log-Polar space that describes the localized shape at each node. We generated weighted adjacency matrices by using normalized histogram differences of subsets of shape context features.
**Dataset and Setup:** We performed an experimental evaluation on 3 datasets. We used a subset of the landmark points provided. The experimental set for each dataset closely follow the setup described in (Caetano et al., 2009). Graph pair instances were generated such that the two graphs are separated by a varying baseline (referred to as "offset" below). Our training data include multiple QAP's for each pair of images. The algorithm learnt a suitable $\omega$. We performed standard 10–fold cross-validation on train/test data for each offset. The number of pairs in the train/test splits varied based on the offset. We summarize our results on various datasets below.
**Hotel/House Dataset:** We considered the CMU house dataset, which contains 111 frames of a video sequence of a toy house. Landmark points were identified and hand–labeled in each frame. Quantitative results corresponding to the matches found on the test set are shown in Fig 4 (left). (Red plots) present the accuracy of matches (on test sets) as the offset (separation between frames) varies. We compared our results against a greedy ("No Learning") matching on two feature settings (blue plots). As expected, no-learn-greedy approach perform poorly if not all features are informative (blue dashed line). For small offsets, no-learn-greedy takes advantage of the fact that the problem instance is easy and the shape context features are useful, but its performance gradually deteriorates as the offset increases (blue bold line). A greedy assignment using learnt weights still performs well. Note that the test phase does *not* perform any backtracking

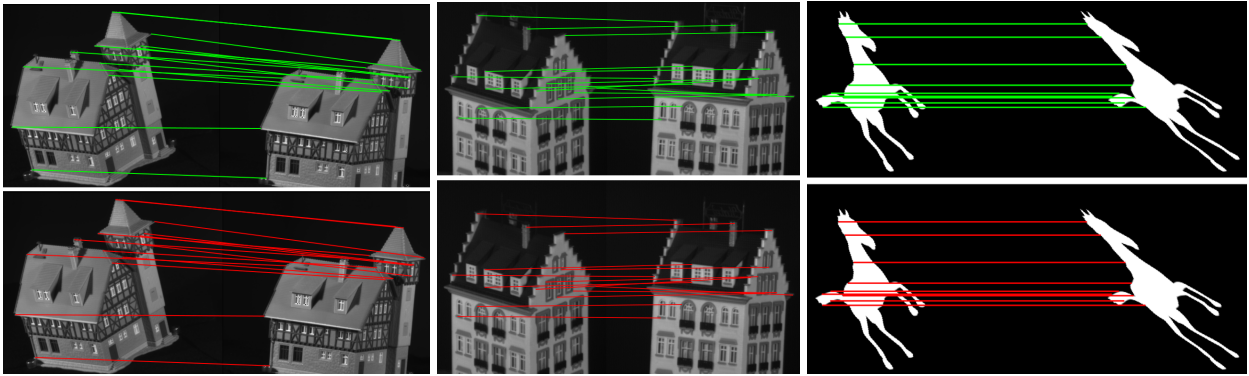

*Figure 3.* A representative pair of permutation is visualized, with the ground truth (green) and the learnt correspondences (red). (Left) House $27 - 97$ frame. (Center) Hotel $58 - 98$ frame. (Right) Shear $66 - 146$ frame.

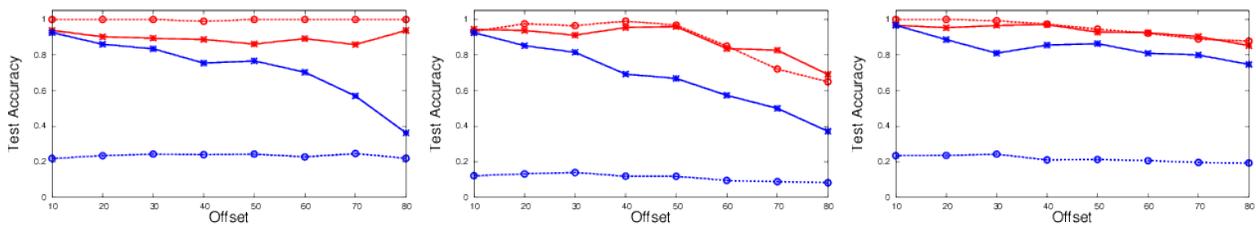

*Figure 4.* Results for our method compared with no-learn baseline. Two colors represent learning (red) and no-learning (blue). (Dashed) Delaunay, distance and 5 uninformative features. (Bold) Delaunay, distance and shape context features. (Left) House, (Center) Hotel, and (Right) Silhouette.

in Branch and Bound.

We performed a similar experiment on CMU hotel dataset, which contains 101 frames of a video sequence of a toy hotel. Again, we see good overall agreement with the ground truth using the learnt parameters. Quantitative results are shown in Fig 4(center). Qualitative results corresponding to the learnt matches on the test set shown in Fig 3

**Silhouette Dataset:** For our second experiment, we used the Silhouette database. We applied horizontal shear to twice its width and transformed the images synthetically. Results for this experiment are shown in Fig 4 (right). Introducing uninformative features makes matching problem more difficult (blue dashed line). While shape context features are quite useful, the learning setting still outperforms no-learning for all offset variations.

Finally, we analysed learnt parameters for the setup that includes Delaunay and distance–based adjacency matrices at 5 scales along with random graph pairs. Learning induced parameters corresponding to Delaunay and distance. Fig 5 shows an example of average weights produced in 10–fold cross-validation setup indicating reduced weights on uninformative features (e.g., non-informative distance scales).

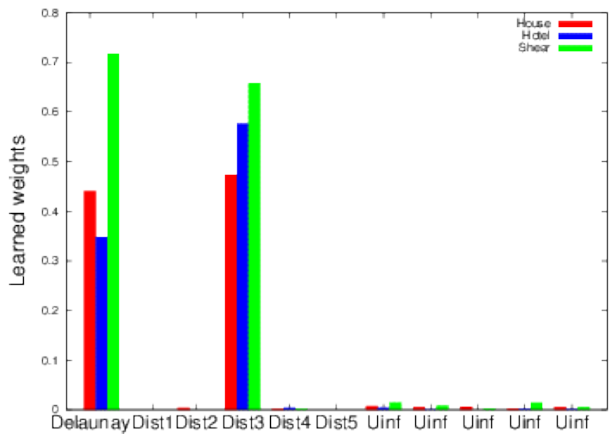

*Figure 5.* Learnt weights for base QAP objective with Delaunay, distance, and uninformative (Uinf) features based QAPs.

## 6. Conclusions

This paper shows that parameter learning (for setting up domain driven compatibility functions) for a general class of hard combinatorial optimization problems can be performed efficiently if the solution to the primary objective function is a member of $\mathbb{S}_n$. We present a framework for performing weight updates on the nodes of a Coset tree. Observe that while the number of leaves of this tree is still $n!$, for the functions discussed here, the bandlimited nature of its Fourier transform makes the process tractable. Our algorithm is inspired by recent results (Kondor, 2008; Huang et al., 2009; Huang & Guestrin, 2010) showing how concepts from harmonic analysis tie to topics in machine learning. Our procedure generalizes to other problems that can be cast as appropriate functions on $\mathbb{S}_n$, and provides a complementary approach to a number of problems typically tackled using Structure Learning. We believe there is additional structure that is not explicitly leveraged by our current model – for instance, we are evaluating the computational benefits of performing weight updates on the frequency components instead of the features. Exploring these properties may provide strategies for other seemingly unrelated problems. For instance, very recently, some related ideas have been independently investigated for submodular set functions (Stobbe & Krause, 2012). Our implementation will be available at http://pages.cs.wisc.edu/~pachauri/.

**Acknowledgments** The authors thank Shamgar Gurevich, Nigel Boston, and Jerry Zhu for various suggestions. This work was supported in part by grants NIH AG034315, NSF RI 1116584, and UW-ICTR (via NIH CTSA award to UW). Collins is supported by the CIBM Training Program (NLM 5T15LM007359).

# References

Belongie, S., Malik, J., and Puzicha, J. Shape matching and object recognition using shape contexts. *PAMI*, 24: 509–522, 2002.

Caelli, T. and Caetano, T. Graphical models for graph matching: Approximate models and optimal algorithms. *Pattern Recognition Letters*, 26(3):339–346, 2005.

Caetano, T., McAuley, J. J., Cheng, L., Le, Q. V., and Smola, A. Learning graph matching. *PAMI*, 31(6):1048–1058, 2009.

Cela, E. *The Quadratic Assignment Problem: Theory and Algorithms*. Kluwer Academic, 1998.

Clausen, M. Fast generalized Fourier transforms. *Theoretical Computer Science*, 67(1):55–63, 1989.

Diaconis, P. Group Representations in Probability and Statistics. *Institute of Mathematical Statistics Monograph Series (Lecture 11)*, 1988.

Finley, T. and Joachims, T. Training structural svms when exact inference is intractable. In *ICML*, 2008.

Hancock, E. and Wilson, R. Graph-based methods for vision: A Yorkist manifesto. *SSSPR*, 2009.

Hartley, R. and Zisserman, A. *Multiple view geometry in computer vision*, volume 2. Cambridge Univ Press, 2000.

Huang, J. and Guestrin, C. Uncovering the riffled independence structure of rankings. *CoRR*, 1006.1328, 2010.

Huang, J., Guestrin, C., and Guibas, L. Fourier theoretic probabilistic inference over permutations. *JMLR*, 10: 997–1070, 2009.

Joachims, T., Finley, T., and Yu, C. N. Cutting-plane training of Structural SVMs. *Machine Learning*, 77(1): 27–59, 2009.

Kondor, R. *Group Theoretical methods in Machine Learning*. PhD thesis, Columbia University, 2008.

Kondor, R. A Fourier space algorithm for solving quadratic assignment problems. In *SODA*, 2010.

Leordeanu, M. and Hebert, M. A spectral technique for correspondence problems using pairwise constraints. In *ICCV*, 2005.

Leordeanu, M. and Hebert, M. Unsupervised learning for graph matching. In *CVPR*, 2009.

Pachauri, D., Hinrichs, C., Chung, M., Johnson, S., and Singh, V. Topology based kernels with application to inference problems in Alzheimer's disease. *TMI*, (99): 1–1, 2011.

Pardalos, P., Rendl, F., and Wolkowicz, H. The quadratic assignment problem: A survey and recent developments. In *DIMACS Series in Discrete Mathematics*. 1994.

Rockmore, D., Kostelec, P., Hordijk, W., and Stadler, P. F. Fast fourier transforms for fitness landscapes. *Appl. and Comp. Harmonic Anal.*, 12(1):57–76, 2002a.

Rockmore, D., Kostelec, P., Hordijk, W., and Stadler, P. F. Fast Fourier transform for fitness landscapes. *Appl. and Comp. Harmonic Anal.*, 12(1):57–76, 2002b.

Rudin, W. *Fourier Analysis on Groups*. Interscience Publishers, 1962.

Shalev-Shwartz, S., Singer, Y., and Srebro, N. Pegasos: Primal Estimated sub-Gradient Solver for SVM. In *ICML*, 2007.

Stobbe, P. and Krause, A. Learning Fourier sparse set functions. In *AISTATS*, 2012.

Taskar, B., Guestrin, C., and Koller, D. Max-margin Markov networks. In *NIPS*, 2003.

Tsochantaridis, I., Joachims, T., Hofmann, T., and Altun, Y. Large margin methods for structured and interdependent output variables. *JMLR*, 6(2):14531484, 2006.

Umeyama, S. An eigendecomposition approach to weighted graph matching problems. *PAMI*, 10(5):695–703, 1988.

Vishwanathan, S. V. N., Borgwardt, K. M., and Schraudolph, N. N. Fast computation of graph kernels. In *NIPS*, 2007.

Xu, Y., Fern, A., and Yoon, S. Discriminative learning of beam-search heuristics for planning. In *IJCAI*, 2007.